\documentclass[letterpaper]{article} 
\usepackage{aaai2026}  
\usepackage{times}  
\usepackage{helvet}  
\usepackage{courier}  
\usepackage[hyphens]{url}  
\usepackage{graphicx} 
\usepackage{natbib}  
\usepackage{caption} 
\usepackage{algorithm}
\usepackage{algorithmic}
\usepackage{booktabs}
\usepackage{newfloat}
\usepackage{listings}
\DeclareCaptionStyle{ruled}{labelfont=normalfont,labelsep=colon,strut=off} 
\floatstyle{ruled}
\newfloat{listing}{tb}{lst}{}
\floatname{listing}{Listing}
\nocopyright 

\usepackage{xcolor}
\usepackage{tikz}
\usetikzlibrary{positioning,calc}
\usetikzlibrary{shapes.geometric, arrows}
\usepackage{verbatim}
\usepackage{fancyvrb}
\usepackage{tcolorbox}
\tcbuselibrary{listings,skins,breakable}
\usepackage{pgfplots}

\newif\ifaddcomments

\newcommand{\commentout}[1]{ }

\newcommand{\roni}[1]{\ifaddcomments{\textcolor{orange}{[Roni: #1]}}\fi}
\newcommand{\yarin}[1]{\ifaddcomments{\textcolor{violet}{[Yarin: #1]}}\fi}

\tcbset{
  mylisting/.style={
    enhanced,
    colback=white,
    colframe=black,
    coltitle=white,
    fonttitle=\bfseries,
    title=#1,
    colbacktitle=blue!60!black, 
    arc=2mm, 
    boxrule=0.6pt,
    listing only,
    listing options={
      basicstyle=\footnotesize\ttfamily,
      breaklines=true
    }
  }
}

\tikzset{every node/.style={font=\Large}}

\title{Toward PDDL Planning Copilot}

\author{
    Yarin Benyamin,
    Argaman Mordoch,
    Shahaf S. Shperberg,
    Roni Stern,
}
\affiliations{
    Ben-Gurion University of the Negev\\

    bnyamin@post.bgu.ac.il,
    mordocha@post.bgu.ac.il,
    shperbsh@bgu.ac.il,
    roni.stern@gmail.com
    
}

\date{August 18, 2025}
\begin{document}

\maketitle

\begin{abstract}
Large Language Models (LLMs) are increasingly being used as autonomous agents capable of performing complicated tasks. However, they lack the ability to perform reliable long-horizon planning on their own. 
This paper bridges this gap by introducing the \emph{Planning Copilot}, a chatbot that integrates multiple planning tools and allows users to invoke them through instructions in natural language. 
The Planning Copilot leverages the Model Context Protocol (MCP), a recently developed standard for connecting LLMs with external tools and systems. 
This approach allows using any LLM that supports MCP without  domain-specific fine-tuning. 
Our Planning Copilot supports common planning tasks such as 
checking the syntax of planning problems, 
selecting an appropriate planner, 
calling it, validating the plan it generates, 
and simulating their execution.  
We empirically evaluate the ability of our Planning Copilot to perform these tasks using three open-source LLMs. 
The results show that the Planning Copilot highly outperforms using the same LLMs without the planning tools. 
We also conducted a limited qualitative comparison of our tool against Chat GPT-5, a very recent commercial LLM. 
Our results shows that our Planning Copilot significantly outperforms GPT-5 despite relying on a much smaller LLM. 
This suggests dedicated planning tools may be an effective way to enable LLMs to perform planning tasks. 

\end{abstract}

\section{Introduction}

Large Language Models (LLMs) like ChatGPT, Gemini, and Grok, can, among other things, generate coherent text,  solve chemistry related problems~\cite{guo2023can}, 
support electronics design~\cite{alvanaki2025sldb}, 
and perform various software engineering tasks~\cite{nam2024using,chen2024chatunitest,achiam2023gpt, huang2022towards}. However, it has been argued that their abilities stem from statistical pattern recognition rather than true understanding~\cite{kambhampati2024can}. 
Indeed, even state-of-the-art LLMs struggle with tasks involving long-horizon planning\cite{huang2025survey, shojaee2025illusion}. 

\emph{Domain-independent automated planning} is a well-established approach for solving long-horizon, goal-oriented, planning problems~\cite{ghallab2004automated}. 
Unlike Reinforcement Learning (RL) methods that depend on interaction with the environment, domain-independent planning operate on a symbolic domain model that specifies the preconditions and effects of each action. These symbolic domains are often defined using the Planning Domain Definition Language (PDDL)~\citep{aeronautiques1998pddl}. 
A planning problem in the PDDL formalism is defined by two components -- a \textit{PDDL domain} and a \textit{PDDL problem}. The \emph{PDDL domain} defines the environment and the agent's capabilities and the \emph{PDDL problem} defines the specific objects on which it is defined as well as the initial and the goal states.

Research on domain-independent planning have produced a diverse set of automated planners over the years. Such planners accept a PDDL domain and problem and produce solutions -- plans -- with formal guarantees of correctness and in some cases optimality. 
Some planners only support domains with discrete state variables~\citep{helmert2006fast,hoffmann2001ff} while others also support numeric state variables\citep{scala2016interval,hoffmann2003metric}, as well as more complex aspects of the planning problem such as time \citep{baier2009heuristic,coles2009temporal} and non-determinism~\cite{muise2024prp}. 
Planning research also developed a variety of supporting tools such VAL, which validates a given plan is executable and achieves the intended goals~\cite{howey2004val}; scripts for parsing PDDLs; and translators facilitate the creation, manipulation, and conversion of planning problems across different syntax and formalisms. 
Together, these tools form a robust planning ecosystem that provides strong theoretical foundations
as well as reliability and predictability of the generated plans.

While automated planning provides powerful guarantees and a rich ecosystem of tools, applying these planning tools directly requires substantial manual effort. Users must choose appropriate planners based on the problem type, and leverage separate tools for validating and analyzing plans. Each stage of the pipeline—syntax checking, planner invocation, plan validation, and execution—typically involves different tools and formats, creating a fragmented workflow. This complexity can hinder adoption, especially for users who lack expertise in planning formalisms or want to interact with the system through natural language. 
A unified framework that integrates these stages into a single, LLM-based interface could therefore drastically reduce this burden and make planning tools more accessible.

In this work, we introduce the \emph{Planning Copilot}\footnote{https://github.com/SPL-BGU/PlanningCopilot}\yarin{The link contains only the framework, without the experimental parts. It will take a bit of time to arrange everything.}---a chatbot that integrates multiple planning tools and allows users to invoke them through plain English instructions. 
To achieve this, we leverage the Model Context Protocol (MCP)~\cite{anthropic2024mcp}, a recently developed standard for connecting language models with external tools and systems. 
MCP enables LLMs to augment their reasoning with real-world actions such as image generation, web search, and API calls, enabling overall much better performance.
The Planning Copilot includes the following capabilities: (1) \textit{automated planner selection}, allowing the LLM to choose between classical and numeric planners based on the domain; 
(2) \textit{syntactic domain validation}, ensuring that PDDL domains conform to expected standards; 
(3) \textit{plan verification}, where the LLM invokes VAL~\citep{howey2004val}-- a well known plan validator -- to validate a given plan; 
and (4) \textit{plan simulation and trace generation}, providing state inspection capabilities to support debugging and analysis of planning outcomes. 
These features allow LLMs to act as effective assistants for AI researchers.
In particular, they complement natural language to PDDL (NL-to-PDDL) translation pipelines by enabling the planning, validation, and execution of generated models.

\section{Preliminaries}
In this section, we provide the necessary background and review previous work relevant to our study. 
We begin by introducing the concept of integrating large language models with external tools.
Then, we give a brief introduction to Automatic Planning and the Planning Domain Definition Language (PDDL)~\cite{aeronautiques1998pddl}. 
Next, we discuss prior work on using LLMs to solve planning problems.

\subsection{LLMs as Agents}

In recent years, numerous large language models (LLMs) have been developed~\citep[inter alia]{team2023gemini,touvron2023llama,gunasekar2023textbooks,liu2024deepseek}. These models enable sophisticated natural language understanding and generation, supporting applications ranging from question answering and summarization to code synthesis and educational tools \cite{lewis2019bart, jiang2024survey, kasneci2023chatgpt}.
Despite their strengths in producing coherent and contextually appropriate text, LLMs natively lack the ability to sense or act within their environments and thus cannot function as autonomous agents on their own.
To overcome these limitations, LLMs have been equipped with \emph{tools} that they can invoke to interact with external environments (e.g., reserving tables in restaurants, booking hotels), thereby enabling them to act autonomously~\cite{zeng2023agenttuning}. 
Early efforts such as Toolformer~\cite{schick2023toolformer} investigate pre-training LLMs to recognize when and how to invoke each tool. 
More recent work on standardized protocols~\cite{ding2025unified}, such as MCP, allows modular, runtime integration, of external tools without any specific fine-tuning. 
Through such tool integration, 
agentic AI systems can perform a wide array of tasks --including image generation, web search, and data analysis -- while staying coherent and aware of the context.

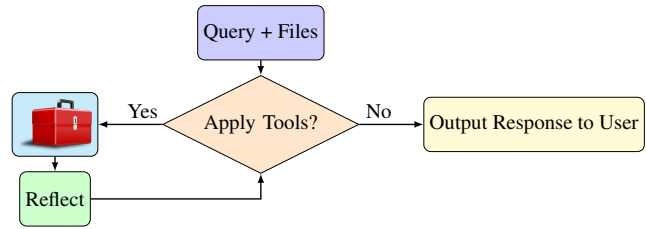
\begin{figure}[ht]
\centering
\resizebox{\columnwidth}{!}{%
\begin{tikzpicture}[
    node distance=0.3cm,
    auto,
    >=latex,
    decision/.style={
        diamond, draw, fill=orange!20, text width=3cm, aspect=2, align=center, inner sep=1pt
    },
    box/.style={
        draw, rounded corners, minimum width=1.1cm, minimum height=1.1cm, align=center, fill=#1!20
    },
    arrow/.style={thick,->}
]

\node[box=blue] (start) {Query + Files};
\node[decision, below=of start] (decide) {Apply Tools?};
\node[box=cyan, left=of decide, xshift=-1.0cm] (tool) {\includegraphics[width=1.5cm]{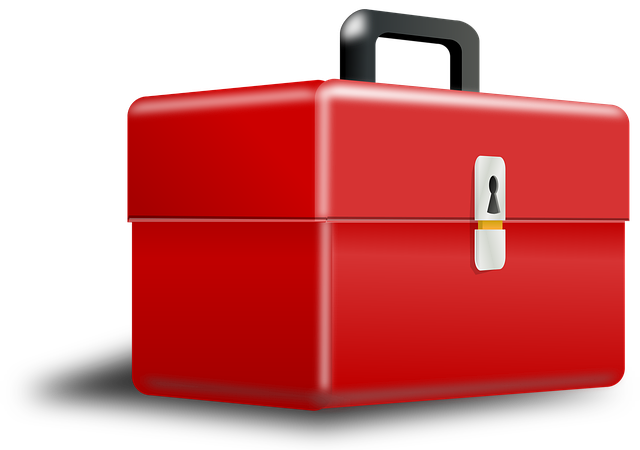}};
\node[box=green, below=of tool] (reflect) {Reflect};
\node[box=yellow, right=of decide, xshift=1.0cm] (output) {Output Response to User};

\draw[arrow] (start) -- (decide);
\draw[arrow] (decide.west) -- ++(-0.8,0) node[midway,above]{Yes} -- (tool.east);
\draw[arrow] (tool) -- (reflect);
\draw[arrow] (reflect.east) -- ++(0,0) -| (decide.south);
\draw[arrow] (decide.east) -- ++(0.8,0) node[midway,above]{No} -- (output.west);

\end{tikzpicture}%
}
\caption{High-level diagram of an LLM using tools.}
\label{fig:state-graph}
\end{figure}

\commentout{
To illustrate how agentic AI works, we can view an agentic system as having two components: the \emph{brain} and the \emph{body}~\cite{agents-course}.
The \emph{brain} is responsible for reasoning and planning—the AI model processes information and decides which actions to take. For this purpose, we use a large language model (LLM).
The \emph{body} is the component that enables the agent to interact with the external world and execute its plans, including through tool use. Methods such as Toolformer~\cite{schick2023toolformer} demonstrate how LLMs can learn to integrate external tools directly into their reasoning process. 
By selecting and leveraging the right tools or knowledge sources for each subproblem, LLM-based agents can tackle complex tasks more effectively. 
Moreover, tool usage provides reliable, execution-based results, rather than relying solely on the language model’s internal knowledge, which may lead to hallucinations.
}
While these tools help LLMs perform complex tasks, they still struggle with reliable long-horizon planning and dynamic goal decomposition, which limits their ability to execute multi-step strategies independently~\cite{huang2025survey, shojaee2025illusion}.

\subsection{Automatic Planning}

Automated Planning is a subfield of Artificial Intelligence (AI) focused on generating plans—which are sequences of actions—that intelligent agents can execute to achieve their goals. 
Automated Planning can be defined through a variety of problem formulations. In this work, we focus on Planning Domain Definition Language (PDDL)~\citep{aeronautiques1998pddl} and its later extension PDDL 2.1~\citep{fox2003pddl2}.
The PDDL (and PDDL 2.1) formalism includes two components---the \textit{domain} and the \textit{problem}. 
The domain describes the environment using predicates and functions, as well as the agent's capabilities, i.e., the actions. 
The problem contains the problem-specific objects, a definition of the initial state from which the planning algorithm begins to find a solution, and the desired goal.
Together, with this precise knowledge, planners such as FastDownward~\citep{helmert2006fast}, Metric-FF~\citep{hoffmann2003metric}, ENHSP~\citep{scala2016interval}, and NYX~\citep{piotrowski2024real} can efficiently generate plans and can offer formal guarantees about their correctness and optimality.
We consider two types of planning: \emph{classical planning}, where actions deterministically change a finite set of propositional states, and \emph{numeric planning}, which extends this model with numeric functions to capture quantities such as resources or costs.

\subsection{Leveraging LLMs for Automated Planning}

\citet{valmeekam2024llms} demonstrated that LLMs struggle with basic planning tasks, while \citet{pallagani2023understanding} showed promise using fine-tuned sequence-to-sequence models like Plansformer~\cite{pallagani2022plansformer}, though with limited generalization to out-of-distribution domains.
The LLM+P framework~\cite{liu2023llm+} converts natural language planning problem descriptions into PDDL using GPT-4, 
then translates the resulting solutions back into natural language for robot planning scenarios. Similarly, GenPlanX~\cite{borrajo2025genplanx} completes a more comprehensive cycle by using LLMs to generate problem instances, which are then processed by a hand-coded pipeline that invokes the planner to solve them. 
\roni{The statement below that says this is a hand-coded pipeline is important. Is it true for LLM+P also? if so, we can say the above methods integrate LLM and planning in a hand-coded dedicated pipeline, while in our case we let the LLM choose when and if to use a planning tool.}
However, significant challenges remain not only in generating accurate PDDL models but also in validating, executing, and debugging these models to ensure practical usability. The Planning Copilot we propose in this work addresses these challenges.

\section{The Planning Copilot}

The Planning Copilot we propose integrates external planning-related tools into a given LLM. 
It allows the LLM to perform long-term planning and validate its outcome, and relieves users from the burden of running, validating, and adapting inputs for these planning tools by providing a natural language interface mediated by the LLMs.

Our Planning Copilot supports three main use cases, denoted here as \textbf{Solve}, \textbf{Validate}, and \textbf{Simulate}. We describe each of these use cases below. 
Each use case is supported by one or more MCP functions. 
These MCP functions allow the LLM to call these planning tools and use the information they output to support the desired use case. 
Notably, this modular design makes it straightforward to integrate with other systems---such as those that generate PDDL domains and problems from natural language---thereby extending our Planning Copilot’s capabilities in the future.

\paragraph{The Solve use case}  
    In this use case, the Planning Copilot is required to solve a planning problem described by a given PDDL domain and problem. It should do so by calling a suitable planner. 
    To support this use case, the Planning Copilot includes two MCP functions, one that calls a classical planner and one that calls a numeric planner. Both calls accepts a PDDL domain and problem specification, executes a planner, and returns the resulting plan to the LLM.
        
    Importantly, the user is not obligated to specify which planner to use. The LLM should infer from from the input files which planning algorithm should be called. 
    This is reasonable because LLMs are trained on data from the web, and different planning domains (classical and numerical) are available in open source repositories. 
    
\paragraph{The Validate use case}  
    In this use case, the Planning Copilot is required to validate either the PDDL domain, the PDDL problem, or the plan generated by some planner. 
    That is, the validation functionality supported by the component includes:  
    \begin{itemize}
        \item Checking the syntactic correctness of the PDDL domain specifications.
        \item Ensuring the PDDL problem is consistent with given PDDL domain.
        \item Validating a given plan is correct, i.e., it achieve the intended goal state according to the given PDDL domain and problem.
    \end{itemize}

    The corresponding MCP function supports validation at various stages of planning: domain definition, problem formulation, and generated plans. 
    If the validation fails, the tool will notify the LLM.     
    In plan validation specifically, the LLM is given the reason for failure and a possible fix suggestion.

\paragraph{The Simulate use case}  
    In this use case, the Planning Copilot is required to simulate the step-wise execution of a given plan. Its MCP function takes a domain, problem, and plan as input and generates a detailed execution trace that records the intermediate state after each action activation.    
    Like the other MCP functions, it allows the LLM to incorporate the verified outputs from the tools back into its context, enabling it to use that information for subsequent reasoning and to provide the user with a more reliable and comprehensive response. For example, given a trace, the LLM could retrieve the 5th state or the 10th action.
    \roni{Not clear if the tool simulates a trace and the LLM needs to find the 5th state of if the tool accepts the state index and does this}

Additionally, a dedicated MCP function saves plans locally to support tool chaining and pipelining, allowing other components or tools to access and continue processing the plan. 
\roni{Not important now, but I don't fully understand this part. Ignore this comment for now. }

\paragraph{Workflow Management}
We implement our Planning Copilot’s control flow using \textit{LangGraph}~\cite{wang2024agent}, an Agentic AI framework that represents workflows as a stateful graph of nodes (LLM calls, tools) and directed edges (control decisions) as seen in Figure~\ref{fig:state-graph}.
Tool selection is handled through a loop between MCP and the LLM: MCP provides the LLM with a list of available tools, including their purpose, inputs, and outputs, and the LLM chooses the one that best matches the current goal. MCP executes the selected tool, records the results in an execution trace, and returns this feedback to the LLM, which reflects on the trace to validate correctness and adjust future selections if needed. 
Figure~\ref{lst:llm_instructions_tools} shows a snippet of the system prompt used to steer tool selection and error-recovery.
\roni{The prompt does not make sense: why are you a PDDL planner - the co pilot does much more than that. It validates, simulates, etc.. Too late to change for now though}

\begin{figure}[t]
\centering
\begin{minipage}{\linewidth}
\begin{lstlisting}[language={}, basicstyle=\ttfamily\small, frame=single, numbers=none]
You are a PDDL planner in a tool-based planning system. 

Your ONLY way to get information or solve problems is by calling the provided tools ONE AT A TIME - never guess or create plan details yourself.

On each turn, either:
- Call the next tool (name + inputs), or
- Give the final answer in natural language BASED ONLY on tool outputs so far.

Do not combine tool calls or answer prematurely. Final answers must contain no tool commands or intermediate data.

Available tools: {tool_list}. Choose the most suitable based on the problem. If a tool returns nothing, treat the problem as unsolvable. Always wait for results - never guess.
\end{lstlisting}
\end{minipage}
\caption{System prompt of our Planning Copilot, which is allowed to call planning tools.\roni{Can make this much smaller}}
\label{lst:llm_instructions_tools}
\end{figure}

\begin{figure}[t]
\centering
\begin{minipage}{\linewidth}
\begin{lstlisting}[language={}, basicstyle=\ttfamily\small, frame=single, numbers=none]
You are a PDDL planner capable of directly reasoning about and generating plans.

You can analyze PDDL problems, validate syntax, create plans, and simulate state transitions all on your own.

Carefully reason step-by-step about problem requirements and constraints before producing outputs.

Do NOT wait for or rely on external tools your responses should be fully self contained.

For tasks: {task name: required format}

Answer only for the specific task you are asked; do not provide information about other tasks.

\end{lstlisting}
\end{minipage}
\caption{System prompt for the baseline LLMs, which are not given the planning tools.\roni{can make this MUCH smaller}}
\label{lst:llm_instructions_no_tools}
\end{figure}

\section{Experimental Results}
We have implemented our Planning Copilot over several open-source LLMs. 
In this section, we provide information about our  implementation and present experimental results comparing the performance of our Planning Copilot with baseline LLMs.  

\subsection{Implementation Details}

For the \texttt{Solve} use case, we created an MCP function that call  FastDownward~\citep{helmert2006fast}, which is a planner for classical planning problems, and an MCP function that calls Metric-FF~\citep{hoffmann2003metric}, a planner for numeric planning problems. 
For the \texttt{Validate} use case, we created an MCP wrapper of VAL~\citep{howey2004val}, a plan validation tool. 
For the \texttt{Simulate} use case, we implemented an MCP function that runs a simple plan simulation script \footnote{\url{https://github.com/argaman-aloni/pddl_plus_parser}}. 

The choice of LLM is flexible; we only require a model that supports MCP. In our implementation, we chose to compare a family of LLMs by Alibaba called Qwen3~\citep{yang2025qwen3} and the GPT-OSS~\footnote{\url{https://openai.com/index/introducing-gpt-oss}} model, both of which support MCP and the so-called reasoning capabilities.
Qwen implemented an “think” mechanisms that allow it multi-step self-reflection, often employing techniques such as the Chain-of-Thought (CoT) mechanism~\cite{wei2022chain}.
Previous results showed that the Qwen3:4B model, even without explicit ``thinking'' mechanisms, \roni{what is this thinking mechanism? can you say more? either here or in the background? why are we talking about it? perhaps explain this mechanism and explain we consider it as it is supposed to produce higher quality solutions}
outperformed other non-thinking models (including LLaMA-3.1-8B-Instruct). When augmented with thinking capabilities, Qwen3:4B achieved competitive performance against larger thinking models (such as DeepSeek-R1-Distill-Qwen-14B), surpassing them in most evaluations. 
More over, GPT-OSS is the latest open-source model, employs a Mixture-of-Experts (MoE)~\cite{shazeer2017outrageously} architecture with a focus on tool-augmented reasoning and agentic workflows, making it an ideal candidate for our PDDL tool-using framework.
\roni{Why did you choose GPT-OSS? if it is the latest, perhaps add here when it was published. This will impress, I believe}

\subsection{Experimental Design}
\roni{Better to say clearly what are the research questions, and then explain that two setups: one where you test a single use case, and one where you test a composition.}

In our experiments, we used ten domains---\textsc{barman}, \textsc{blocksword}, \textsc{depots-classical}, \textsc{rovers}, \textsc{satellite}, \textsc{depots-numeric}, \textsc{counters}, \textsc{farmland}, \textsc{sailing}, and \textsc{Minecraft-Pogo}.
The first five domains are from the the classical track in International Planning Competition (IPC) \citep{long20033rd}, 
the next four are from the numeric IPC track 2023~\cite{taitler20242023}, and the last one is non-IPC numeric domain~\cite{benyamin2024crafting}.

We evaluate our system in a zero-shot setting, i.e., without any dedicated pre-training. 
First, we test whether the use of tools improves the LLM's performance in answering the tasks. In Figure~\ref{lst:llm_instructions_no_tools} we can see the system prompt used when the LLM does not have access to tools.
Second, in the tool-using condition, \roni{what is the "tool-using condition"?}
we measure the percentage of correct task answers. 
The first evaluation compares LLM performance with and without tool usage, while the second considers only runs with tools.
Note that in the results we evaluated two versions of each model—with and without the model's internal deliberation (``thinking'') ability—and reported the better of the two.\roni{Confusing. Say in one place (background?) what is ``internal deliberation'' and then call it either ``internal deliberation'' or ``thinking''}
\roni{This is confusing. Would've been better to keep on option consistently.}

Each task in our evaluation consists of 10 distinct problems, including 5 numeric problems and 5 classic problems. For each problem type, we generate 5 variations of the task prompt to test model robustness across slightly different formulations. 
\roni{show an example of these prompts}
The first evaluation contains five core tasks involving a single request that can be easily solved by an expert using one of the provided tools. 
\roni{not sure what is this "easily" and "expert" here. }
These tasks include: (1) given a domain, check its syntax; (2+3) given a domain and problem, check the problem’s syntax or find a solution; and (4+5) given a domain, problem, and plan, verify that the plan solves the problem or simulate its execution. 
\roni{Not sure what are these numbers -- 2+3, 4+5? all this is not organized well.}

When we have simulation task we can also request returning a specific states during its execution. To capture this, we define tasks that can be chained together, where the number of subtasks in the chain is denoted as $no.tasks$. A task chain of length $no.tasks$ may require up to $no.tasks - 1$ tool calls to solve, with the difficulty increasing as the chain length grows. The evaluation includes tasks across different chain lengths: 250 tasks for $no.tasks = 1$ (derived from 10 problems $\times$ 5 prompt variants $\times$ 5 different requests), 100 tasks each for $no.tasks = 2, 3, 4$, and 50 tasks for $no.tasks = 5$, resulting in a total of 12 task chains of varying complexity for the second evaluation.

We also have a task that request returning specific states within the transition, this task is depend on running the simulation correctly.
By chain one task to another we obtain a total of 12 tasks of varying $no.tasks$ (number of task that chain toghether) for the second evaluation.
Each of these tasks can require $no.tasks - 1$ tool calling to solve the tasks, with a maximum of $no.tasks - 1$ tools per level. 
\roni{what is no.tasks?}
There are 250 tasks in $no.tasks=1$ (10 problems, 5 variants per prompt, and 5 different requests), 100 tasks for levels $no.tasks=2,3,4$, and 50 tasks for the $no.tasks=5$.
\roni{Sorry. All this not clear.}

To assess model performance, we performed validation on the plan solutions using VAL. The validation evaluated whether the answers were correct, and the simulated outputs were compared to the real simulation results using the capabilities of pddl\_plus\_parser ability.
The outputs then were manually evaluated to verify whether the final results provided to the end user were correct. This ensured that both correctness and applicability of the generated plans were taken into account.
When evaluating tool usage, we only consider cases where the LLM explicitly employed the correct tools needed to accomplish the task,
preventing unverified answers from being returned to the user.

Experiments were conducted via API on LangGraph while all code, prompts, temperature, and evaluation scripts were fixed prior to testing to prevent unintended modifications and ensure consistency across experiments.
Specifically, the experiments ran on an RTX 6000 GPU with 48 GB of VRAM. 
Each model used the same system prompt and tool settings, with the randomness parameter (\texttt{temperature}) set to 0. \roni{but some models did not have tools, no?}
Self-loops within the LangGraph were limited to 10 iterations. \roni{where did you use self-loops? why do you need self-loops?}

\yarin{
In this section we first disscuss the first evaluation using the 5 core task, and then the second one that contain ..}

\roni{In all results, give also examples: "... outperformed all other ... For example, in task X model Y did Z and model Y+tools did Z+"}

\subsection{Results: Single Task Comparison}

Table~\ref{tab:no-tool-results} presents the results of our evaluation comparing LLMs with and without access to planning tools. 
Each row represents a different model 
, and the columns correspond to the different tasks we have: Solve, Validate Domain, Validate Problem, Validate Plan, and Simulate.

As expected, LLMs augmented with these tools outperformed those without them. 
As an example, without using tools, none of the LLMs could simulate the plan, while Qwen-3:0.6B succeeded 6\% of the time, Qwen-3:4B succeeded 24\%, and GPT-OSS:20B succeeded 80\%. 
Specifically, except for the problem validation task, GPT-OSS augmented with the tools outperformed all other models.
The task that was the most challenging for LLMs without planning tools was the simulation task, as it requires the model to track state changes and manage the add and delete effects of each action across multiple steps. In contrast, the different validation tasks was executed most successfully among all the tasks, though it can occasionally be answered correctly by chance.
This highlights the importance of planning tools: they ensure that the plans returned by the LLM are not only applicable but also verified through real validation or simulation. Tools like those in our system provide users with confidence that the results they receive are accurate and reliable.

\begin{figure}[t]
    \centering
    \includegraphics[width=0.65\linewidth]{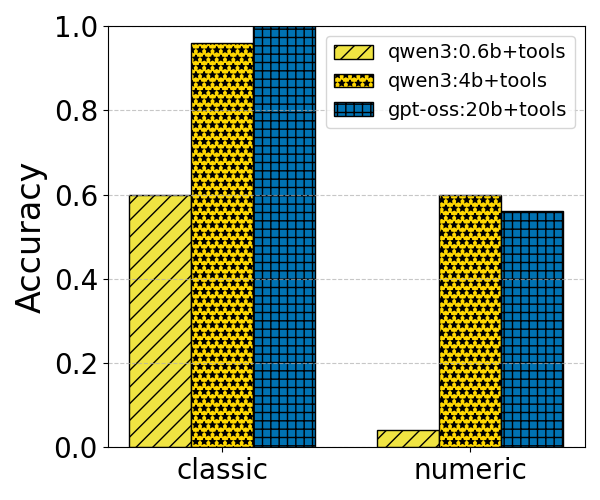}
    \caption{Accuracy of choosing the correct planner for different types of planning problems (classical or numeric).
    }
    \label{fig:planner-choose}
\end{figure}

After establishing that tools improve LLM solving performance, we now focus on how well these models select the appropriate tool for the Solve task. In this task, the LLM receives a domain and a problem and must select a planner to generate a solution. Figure~\ref{fig:planner-choose} shows a subset of the results, highlighting how often the LLM chose the correct tool: a classical planner for classical PDDL problems and a numeric planner for numeric PDDL problems. The results indicate that the LLM was able to solve almost all classical planning problems using the correct tool, while numeric planning proved more challenging. This may be because numeric planning is less common and more complex, so the model likely has less exposure to numeric problems in its training set.

\subsection{Results: Multi-Task Comparison}

\begin{figure}[t]
    \centering
    \includegraphics[width=0.85\linewidth]{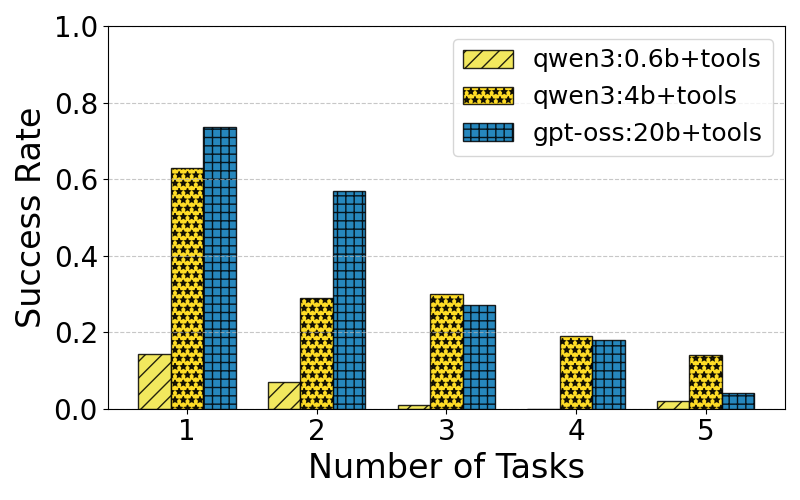}
    \caption{Accuracy of task completion plotted against the number of distinct tasks in a single prompt.
    }
    \label{fig:task-results-man}
\end{figure}

Figure~\ref{fig:task-results-man} presents the performance of tool using of LLMs 
on more challenging tasks that require multiple tool calls in sequence to fulfill the user’s request. 
For instance, tasks that require a single tool run, as shown in 
Figure~\ref{tab:no-tool-results} are averaged and represented as a single bar at the first position.
As expected, we observe that performance declines across different LLMs as the number of required tool calls increases.
Notably, GPT-OSS achieves the best results on tasks involving two or fewer tool calls.
This highlights the importance of robust multi-step reasoning capabilities in LLMs, as errors can compound across sequential tool executions, significantly impacting overall task performance.  
Future work could explore strategies for decomposing large prompts into smaller, manageable subtasks, improving error detection and recovery, and dynamically selecting tools based on intermediate outputs to better handle complex, multi-request prompts.

\begin{table}
\resizebox{\columnwidth}{!}{
\begin{tabular}{@{}lccccc@{}}
\toprule
Model             & Solve & Val. dom. & Val. prob. & Val. plan & Simulate \\ \midrule
qwen3:0.6b        & 0.00  & 0.00      & 0.00       & 0.00      & 0.00     \\
qwen3:0.6b+tools  & 0.32  & 0.14      & 0.16       & 0.04      & 0.06     \\
\midrule
qwen3:4b          & 0.04  & 0.42      & 0.20       & 0.14      & 0.00     \\
qwen3:4b+tools    & 0.78  & 0.34      & 0.98       & 0.80      & 0.24     \\
\midrule
gpt-oss:20b       & 0.28  & 0.20      & 0.14       & 0.54      & 0.00     \\
gpt-oss:20b+tools & 0.78  & 0.70      & 0.40       & 1.00      & 0.80     \\ \bottomrule
\end{tabular}
}
\caption{Success rate of single-task completion shown by task description. We denote the validation task as Val, the domain as dom, and the problem as prob.}
\label{tab:no-tool-results}
\end{table}

\subsection{Comparison with GPT-5}
We performed a small-scale qualitative comparison with the state-of-the-art LLM GPT-5\footnote{GPT-5 is a state-of-the-art model published world-wide on August 7th 2025.}. Specifically, we executed queries using the chat itself, and we did not run the code using the API.  
We allowed the chat to think longer to provide with the best response it could, i.e., we enabled the LLM to select (if needed) to use the GPT-5 Thinking model, which may think longer to provide better response.
For each query we opened a new chat to enable each query to run independently.
We performed the following tasks: (1) solve, (2) domain validation (3) problem validation, (4) plan validation and (5) plan simulation. 
Each task was preformed on 5 different instances.
We used the same queries as used for the models without the tools.
As access to the GPT-5's API is costly, we selected to perform a small preliminary evaluation to compare our approach against this flagship model. 

\begin{figure}[t]
    \centering
    \includegraphics[width=0.9\linewidth]{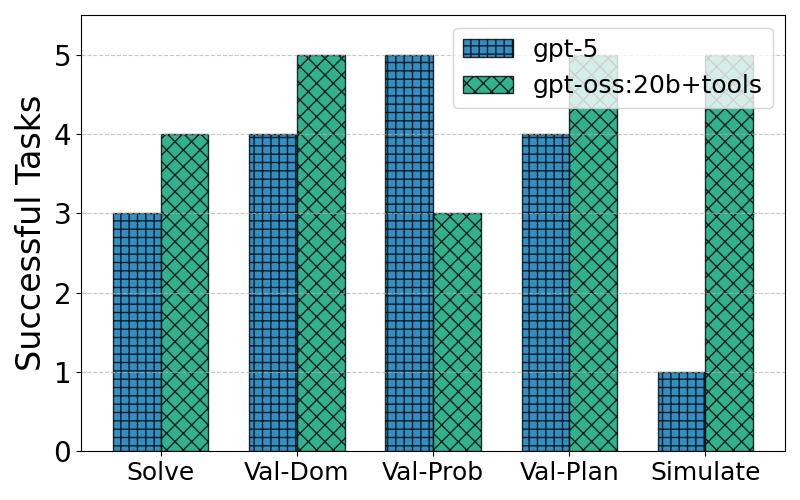}
    \caption{Preliminary exploration of ChatGPT 5 performance compared to the significantly smaller GPT-OSS:20b augmented with our suite of tools.}
    \label{fig:gpt5-results}
\end{figure}

Figure~\ref{fig:gpt5-results} presents the comparison results between our tool, based on the GPT-OSS model, and GPT-5. The x-axis represents the different tasks given to the model, the y-axis represents the number of tasks completed by the model. The green columns represent the results for our tool and the blue columns represent the results for GPT-5.

For the solving task, GPT-5 generated three valid plans and failed in solving two tasks. Specifically, we observed that it successfully solved the problems for the \textsc{Barman}, \textsc{Blocksworld}, and \textsc{Minecraft} domain, but failed to solve the problems for the \textsc{Rovers} and \textsc{Sailing} domains.
Our tool however, succeeded in solving all planning tasks except for the \textsc{Minecraft} domain. In this domain the tool did not understand which planning algorithm it should execute, thus failed.

GPT-5 coded a parser to parse the planning domain and problem, and using its reasoning it attempted to generate the actions that provide solution for the problem.
The LLM understood that it needs to consider the action's preconditions and it even simulated its generated plans internally to validate their correctness.
Subsequently, using this approach it solved three out of the five tested problems. 
Contrary to GPT-5, using a significantly smaller LLM --- GPT-OSS, without the ability to code but augmenting with planning algorithms, we were able to solve more problems by providing it with our planning copilot tools.

GPT-5 performed the best in the validations tasks with at least 4 out of five tasks successfully completed. This is to be expected as this task only involves syntactic understanding of the text that can be achieved by building an Abstract Syntax Tree (AST). 
Our tool outperformed GPT-5 in the plan validation --- the most difficult validation task, and domain validation tasks with all instances successfully validated.
Surprisingly GPT-5 performed better in the problem validation task. We observed that the failed instances belonged to the \textsc{Minecraft} and \textsc{Sailing} domains. We observe that in general, our tool had more difficulties correctly resolving \textsc{Minecraft} related tasks. For the \textsc{Sailing} instance, the tool's context became corrupted due to internal reasoning issues and it failed to perform the validation.

Finally, the least successful task for GPT-5 was the Simulate task with the model only successfully simulating a single instance, specifically the \textsc{Blocksworld} instance.
Our tool, however, successfully simulated all planning instances. 
For example, in instances from the \textsc{Barman}, \textsc{Sailing}, and \textsc{Minecraft} domains, we observed that GPT-5 chose to output only part of the simulated trajectory. 
For the \textsc{Rovers} instance however, the simulation included all actions but the states were partial, containing a single state variable only.

\section{Conclusions and Future Work}

In this work we presented the Planning Copilot, an LLM tool augmented with AI planning tools.
The planning copilot van solve planning problems, perform validation and simulate plans.
We implemented the tool using three open-source models and compared them to their un-augmented counterparts. 
The results show that that augmenting the LLMs with the planning tools highly outperforms using only the LLMs. We also compared our tool to GPT-5, the state-of-the-art LLM model. 
The comparison showed that in all but one task, our tool, using much smaller model, outperformed GPT-5. 

One promising direction is using an interactive plan visualization tool that could automatically generate intuitive diagrams of the proposed plan, helping human users quickly understand, validate, and refine the agent’s strategy.
Similarly, an automated module for generating PDDL domains from free-text descriptions could significantly reduce the effort required for domain modeling.
By integrating these capabilities with the other tools presented in this paper, we could enable a seamless pipeline from natural language task descriptions to executable symbolic plans, empowering users to move from intent to validated plan more efficiently and transparently.

\bibliography{aaai2026}

\end{document}